\RequirePackage{booktabs}

\documentclass[bst/sn-mathphys-num]{sn-jnl}

\usepackage{graphicx}%
\usepackage{multirow}%
\usepackage{amsmath,amssymb,amsfonts}%
\usepackage{amsthm}%
\usepackage{mathrsfs}%
\usepackage[title]{appendix}%
\usepackage{textcomp}%
\usepackage{manyfoot}%
\usepackage{booktabs}%
\usepackage{algorithm}%
\usepackage{algorithmicx}%
\usepackage{algpseudocode}%
\usepackage{listings}%

\usepackage{my_package}

\theoremstyle{thmstyleone}%
\newtheorem{theorem}{Theorem}
\newtheorem{proposition}[theorem]{Proposition}%

\theoremstyle{thmstyletwo}%
\newtheorem{remark}{Remark}%

\theoremstyle{thmstylethree}%
\newtheorem{definition}{Definition}%

\newtheorem{lemma}[theorem]{Lemma}

\newtheorem{corollary}[theorem]{Corollary}

\begin{document}

\title[Article Title]{Toward Finding Strong Pareto Optimal Policies in Multi-Agent Reinforcement Learning}


\author[1]{\fnm{Bang Giang} \sur{Le}}\email{giangbang@vnu.edu.vn}

\author*[1]{\fnm{Viet Cuong} \sur{Ta}}\email{cuongtv@vnu.edu.vn}

\affil[1]{\orgdiv{Human Machine Interaction Laboratory}, \orgname{VNU University of Engineering and Technology}, \city{Hanoi}, \postcode{10000}, \country{Vietnam}}


\abstract{In this work, we study the problem of finding Pareto optimal policies in multi-agent reinforcement learning problems with  cooperative reward structures. We show that any algorithm where each agent only optimizes their reward is subject to suboptimal convergence. Therefore, to achieve Pareto optimality, agents have to act altruistically by considering the rewards of others. This observation bridges the multi-objective optimization framework and multi-agent reinforcement learning together. We first propose a framework for applying the Multiple Gradient Descent algorithm (MGDA) for learning in multi-agent settings. We further show that standard MGDA is subjected to weak Pareto convergence, a problem that is often overlooked in other learning settings but is prevalent in multi-agent reinforcement learning. To mitigate this issue, we propose MGDA++, an improvement of the existing algorithm to handle the weakly optimal convergence of MGDA properly. Theoretically, we prove that MGDA++ converges to strong Pareto optimal solutions in convex, smooth bi-objective problems. We further demonstrate the superiority of our MGDA++ in cooperative settings in the Gridworld benchmark. The results highlight that our proposed method can converge efficiently and outperform the other methods in terms of the optimality of the convergent policies. {The source code is available at} \url{https://github.com/giangbang/Strong-Pareto-MARL}.}




\maketitle

\section{Introduction}

Multiagent reinforcement learning (MARL) is the learning framework where multiple different agents learn to make optimal decisions in an environment through the Reinforcement Learning paradigm. MARL research advances rapidly, with the recent development of MARL achieving impressive results on a wide range of learning scenarios, notably in zero-sum games \citep{silver2017mastering} and fully cooperative environments \citep{rashid2020monotonic, yu2022surprising} where all agents share the same reward function. 

A less studied area of research in MARL is in the environments with cooperative reward structures, in which all agents have different and possibly conflicting reward schemes \citep{lowe2017multi}. In these problems, there might be multiple different but optimal policies, in the sense that none of which is better than the other. This concept is known as the Pareto optimality in multi-objective optimization. It is desirable that we can find such Pareto optimal policies, as they are equivalent to the optimal policies in the common setting of single agent problems \citep{christianos2022pareto}. While the current MARL methods are known to find Nash equilibrium \citep{leonardos2021global}, such solutions can be suboptimal \citep{papoudakis2020benchmarking}. 

In this paper, we first show that in general cooperative environments, agents need to explicitly consider the optimization of other agents to achieve Pareto optimality. Such behaviors are known as altruistic learning in RL literature \citep{franzmeyer2022learning};  altruistic learners learn to act for the benefit of other agents even though those actions do not bring about any personal gain for the actors. To learn altruistically, one agent needs to optimize not only the reward of itself but also the rewards of other agents, which involves some form of multi-objective optimization. As a result, we connect the multi-objective framework to the MARL domain.

Multiple Gradient Descent Algorithm (MGDA) \citep{desideri2012multiple} is one of the most popular gradient-based multi-objective methods. MGDA can find arbitrary solutions in the Pareto optimal Set using first-order gradient descent. However, MGDA is known to only converge to weak Pareto optimal solutions \citep{fliege2019complexity}. While this problem is not significant in other learning settings, we show that MARL problems can have many weak Pareto Stationary points, which can reduce the efficacy of MGDA. 
To this end, we identify the effect of diminishing gradient norms as a root cause to the weak Pareto convergence issue of MGDA and propose an improved version of MGDA++ based on this observation. We demonstrate both theoretically and empirically that MGDA++ can converge to strong Pareto optimal solutions. To summarize, our contributions in this paper are:
\begin{itemize}
    \item We show that to achieve Pareto optimality in MARL, agents need to consider the objective of other agents, we connect the multi-objective optimization with MARL problems, and propose to apply MGDA to the MARL problems. 
    \item We propose MGDA++, an extension of MGDA that converges to strong Pareto Solutions with bi-objective problems in convex, smooth settings both theoretically and empirically. To our knowledge, this strong Pareto convergence result in the convex setting with gradient descent is the first in the literature. 
    \item We demonstrate the effectiveness of MGDA++ with trust region methods through several cooperative scenarios in the Gridworld benchmark. Our proposed method is able to achieve better convergence solutions across different agents in comparison with other baselines.
\end{itemize}


\section{Background and Motivations}
\subsection{Problem settings of MARL}


A Markov game is determined by a tuple $\langle \mathcal N, \mathcal S, {{\mathcal A}}, P, r, \gamma, \rho \rangle$. We denote $\mathcal N$ the set of all $N$ agents, $\mathcal S$ the state space, ${{\mathcal A}}$ the joint action space of each $i$ agent's action space $\mathcal A_i$, i.e ${{\mathcal A}} = (\mathcal A_1 \times \dots \times \mathcal A_{N})$,
$\rho$ the initial state distribution,
and $\gamma \in [0, 1)$ the discount factor. 
In the cooperative setting, all the agents have a different reward function $r^i:\mathcal{S}\times \mathbf{\mathcal A} \rightarrow \mathbb R$. 
At each timestep $t$, all agents take actions $\mathbf a_t = (a^1_t, a_t^2, \dots, a^{N}_t)$ according to their respective policy $\pi^i$, then the environment transitions to the next state $s_{t+1}$ according to the dynamic kernel $P$. The total expected discounted cumulative returns of each agent are defined as $J_i(\boldsymbol  {\pi}) = \mathbb E_{\boldsymbol  {\pi}} \sum_t \gamma^t r_t^i$. 
Unlike the previous works in the {cooperative} settings \citep{lowe2017multi}, we do not assume each agent to necessarily maximize its returns but to find equilibriums such that no better joint policies exist, such equilibriums are called Pareto optimal solutions, formally defined in the next section.

In multiagent games, two solution concepts prevail: Pareto optimality (PO) and Nash Equilibrium (NE). 
In NE, each agent assumes that the strategies of all other agents remain unchanged, leading to an equilibrium where no agent has the incentive to alter their course of action. 
Conversely, PO involves the notion of collective welfare, wherein each agent endeavors to attain outcomes beneficial to all. PO states are those where no individual agent can increase their gains without diminishing those of others.

\subsection{Problem settings of multi-objective optimization}

\textbf{Notations.} We use $[n]$ to denote the set of the first $n$ positive intergers, $[n] = \{1, \dots, n\}$, all the norms $\|\cdot\|$ used in this paper is assumed to be L2 norm and the inner product $\langle\cdot, \cdot\rangle$ is the dot product. 
Also, $|S|$ denotes the cardinality of a finite set $S$. For two vectors $x$ and $y$, we define $x^i$ the $i^{th}$ element of the vector $x$, and the partial order $x \preceq y$ and $x \prec y$ denote the elementwise inequality $x^i \leq y^i$ and $x^i < y^i$, respectively.

We consider the optimization of the multi-objective function $F: \mathbb R^m \rightarrow \mathbb R^n$ as
\[\min_{x\in \mathbb R^m} F(x) = [F_1(x), F_2(x), \dots, F_n(x)].\]

In contrast to single-objective optimization, solving the Multi-Objective Optimization problem requires making trade-offs between objectives, as it's typically impossible for all objectives to be optimal simultaneously. Similar to single-objective optimization, the concept of stationarity extends to multi-objective settings.
\begin{definition} [Pareto stationary] $x^*$ is a Pareto stationary point if 
\[\text{range}(JF(x^*))\cap -\mathbb R^n_{++} = \emptyset\]
with $JF(x)$ is the Jacobian matrix of $F(\cdot)$ at $x$ and $\mathbb R_{++}$ is the set of positive numbers.
\end{definition}

Similarly, the concept of Pareto optimality is analogous to the global optimality in the single-objective domain.

\begin{definition}\label{def:strongpareto}
    $x^*$ is called a Pareto (optimal) solution if there does not exist any other $x$ such that $F(x)\preceq F(x^*)$ and $F(x)\neq F(x^*)$.
\end{definition}
\begin{definition} \label{def:weakpareto}
    $x^*$ is called a weak Pareto (optimal) solution if there does not exist any other $x$ such that $F(x)\prec F(x^*)$.
\end{definition}
\begin{definition}
    $x^*$ is called an $\varepsilon$-Pareto (optimal) solution if there exists a Pareto optimal solution $x$ such that $ F(x^*) \preceq F(x)+\varepsilon$.
\end{definition}

The Definition of Pareto optimality in Def \ref{def:strongpareto} is sometimes known as the strong Pareto optimality to distinguish it from the weak Pareto one defined in Def \ref{def:weakpareto}. Weak and strong Pareto points are equivalent under certain conditions from the objective function $F$, e.g. strong quasiconvex and continuity \citep{cato2023weak}. In this work, we consider the convex objective functions where weak and strong PO solutions are different (Assumption \ref{ast:cvx_smt}). 
Weak PO can be arbitrarily worse than strong Pareto solutions.

In steepest descent method \citep{fliege2000steepest}, the objective is to find a common update vector $d$ such that it minimizes all the objectives' first-order approximation with an L2 norm regularization, 
\begin{equation}\label{eq:mgda_prob}
    \min_d \max_{i=1,\dots,n} \langle \nabla F_i(x), d\rangle +\frac{1}{2}\|d\|^2.
\end{equation}
This problem can be formulated as a linear program as 
\begin{equation}\label{linp:steepest}
\begin{array}{ll@{}ll}
\text{minimize}  & \displaystyle\alpha + \frac{1}{2}\|d\|^2&\\
\text{s.t.}&  \displaystyle\langle \nabla F_i(x), d\rangle -\alpha \leq 0, &\quad \forall i=1, \dots, n.
\end{array}
\end{equation}
The dual problem of (\ref{linp:steepest}) is the following problem, 
\begin{equation}\label{linp:mgda}
\begin{array}{ll@{}ll}
\text{minimize}  & \left\| \sum_{i}^n\lambda_i \nabla F_i(x)\right\|^2&\\
\text{s.t.}&  \sum_i\lambda_i = 1 \\
& \lambda_i \geq 0&\forall i=1,\dots,n.
\end{array}
\end{equation}
The formulation in (\ref{linp:mgda}) is considered in \cite{desideri2012multiple}, where it gets the name MGDA, and is further developed by \cite{sener2018multi} to work with large-scale neural networks. 
MGDA is known to converge to arbitrary Pareto stationary points.  
Geometrically, the unique solution in (\ref{linp:mgda}) is the smallest norm vector in the convex hull of all the gradient vectors.


\subsection{The difficulty of finding the Pareto optimality with current MARL framework}


One of the challenges in cooperative multiagent reinforcement learning with policy gradients (PG) is that the solutions found by these algorithms are usually suboptimal. 
For example, \cite{christianos2022pareto} provides a matrix game example where the PG converges to a Pareto-dominated policy. Similar results are also reported empirically in \cite{papoudakis2020benchmarking}, who showed that many recent MARL algorithms are unable to converge to the Pareto-optimal equilibrium. 

Theoretical insights by \cite{leonardos2021global} establish that PG methods converge toward NE in Markov Potential Games.
The convergence relies on the independent learning assumption, which states that every agent learns, and acts, using only their local information. 
This suffices for NE, since agents are perceived as static from others' perspectives, rendering local information adequate. On the other hand, \cite{zhao2023local} shows that MAPPO converges to globally optimal policies in fully cooperative MARL, given that agents can observe their peers' actions. 
Without action-sharing, PG methods are only able to converge to NE solutions \citep{kuba2021trust}. The action-sharing mechanism can be seen as a type of communication, in which agents share their strategy and planning. Given their close connections, a natural question arises:

\begin{displayquote}
    \textit{Can policy gradient methods converge to Pareto optimal policies in {cooperative} games, given arbitrary communication?}
\end{displayquote}
Unfortunately, the answer to this question is negative; in certain instances, PG methods fail to converge to PO, even when agents freely exchange their actions. The matrix game example in Figure \ref{fig:game} illustrates this argument. 
\begin{wrapfigure}{r}{0.45\textwidth}
\begin{game}{2}{2}[Player~1][Player~2]
& $A$ & $B$ \\
$A$ &$1,2$ &$0,2$ \\
$B$ &$1,0$ &$0,0$
\end{game}\hspace*{\fill}%
\caption[]{Example of the matrix game, the tuple in each table cell contains the reward of agent 1 and 2, respectively. There is one Pareto Optimal solution at $(A, A)$. Note that in this example, \textit{any} policy profile, stochastic or deterministic, is a Nash Equilibrium.}
\label{fig:game}
\end{wrapfigure}

Given a matrix game in a normal form as in Figure \ref{fig:game}, where each agent's actions solely influence the rewards of the other. 
As a result, any policy profile in this matrix game is a Nash Equilibrium. To see why PG does not converge to PO, we first find the gradient of an agent by using the Policy Gradient theorem \citep{sutton2018reinforcement},
\[\nabla_{\pi^i} J_i = \sum_s d(s)\sum_a A^{\pi^i}(s, a)\nabla \pi^i,\]
{where $d(s)$ denotes the unnormalized discounted marginal state distribution}.
But since the reward of the agent $i$ does not depend on its actions, the advantage function $A^{\pi^i}$ becomes constant, as a result
\[\nabla_{\pi^i} J_i = \sum_s d(s) A^{\pi^i}\sum_a\nabla \pi^i=0.\]
This is logical, because if an agent has no control over the reward they receive - a constant function of action, then the gradient - the rate of change of the return over the actions - of a constant function is zero. 
Therefore, in such situations, Policy Gradient fails to learn.

Note that knowing the action that the other agent will take has no benefit in this case, again because the agent cannot utilize this information to improve its decision which does not affect its reward; an agent has no way to dictate other agents to act upon their own interest.
Furthermore, a similar argument applies to the sharing parameter setting. Because the gradient is zero, no learning happens.

Moreover, we can extend the above observation to value-based algorithms such as IQL and MADDPG \citep{lowe2017multi}, when the policy of one agent is stable, the learned values of an agent is a constant function, and if the choice of the optimal action in Q function to break ties is arbitrary, then the algorithm is also prone to converging to suboptimal policies. We formalize the above idea to a broad class of MARL algorithms as follows.

\textit{Any MARL algorithm where each agent only optimizes their objective is susceptible to converging to Pareto Dominated policies.}

The problem is that each agent selfishly optimizes only their own reward, so no learning can happen if their actions and rewards are uncorrelated. 
This issue can be fixed straightforwardly:
during training, agents should transition from self-centered reward optimization to considering the objectives of other agents as well.
As a result, we have connected the multi-objective ideas to the training of multi-agent RL in cooperative settings. 
Furthermore, we note that although the example presented in Figure \ref{fig:game} seems contrive, more complex scenarios involve the interaction of agents where one agent can help other agents to improve their performance, while greedy agents optimizing only their objective can bring worse outcomes to the whole.  



\section{Multi-objective methods in MARL}\label{sec:3}
\subsection{MGDA with Trust region Policy Optimization}\label{sec:3.1}
In this section, we discuss how MAPPO \citep{yu2022surprising} can be combined with the MGDA algorithm to find Pareto optimal policies. MAPPO is an algorithm designed for fully cooperative settings, where all agents share a common reward function. As a result, for MAPPO to work in multi-reward settings, we first extend the value function to have a multi-head structure, where each head is responsible for predicting the baseline for the corresponding agent. This structure is akin to the multi-head critic in multitask learning \citep{yu2020gradient}. As a result, the objective for optimizing the policy $i$ with respect to the reward $j$ with PPO is as follows
\[L^{i, j}_\text{PPO} = \min\left(\frac{\pi^i_\text{new}}{\pi^i_\text{old}}\hat{A}_j^{\pi^i_\text{old}}, \text{clip}\left(\frac{\pi^i_\text{new}}{\pi^i_\text{old}}, 1-\epsilon, 1+\epsilon\right) \hat{A}_j^{\pi^i_\text{old}} \right),\]
where $\hat{A}_j^{\pi^i_\text{old}}$ is the estimated advantage of the policy $i$ to the reward of agent $j$, with the baselines from the multi-head value function for variance reduction. As a result, for the agent $i$, the set of gradient $\{\nabla L^{i, j}_\text{PPO}\}_{j=1}^n$ w.r.t. other agents' reward can be passed into the MGDA problem in (\ref{linp:mgda}) (and similarly MGDA++ presented in the next section) to find a descent direction that improves on the returns of all agents. The algorithm can then proceed with the usual descent methods \citep{kingma2014adam}.

\subsection{The problem of MGDA}
\begin{wrapfigure}{r}{0.4\textwidth}
\centering
\includegraphics[width=0.39\textwidth]{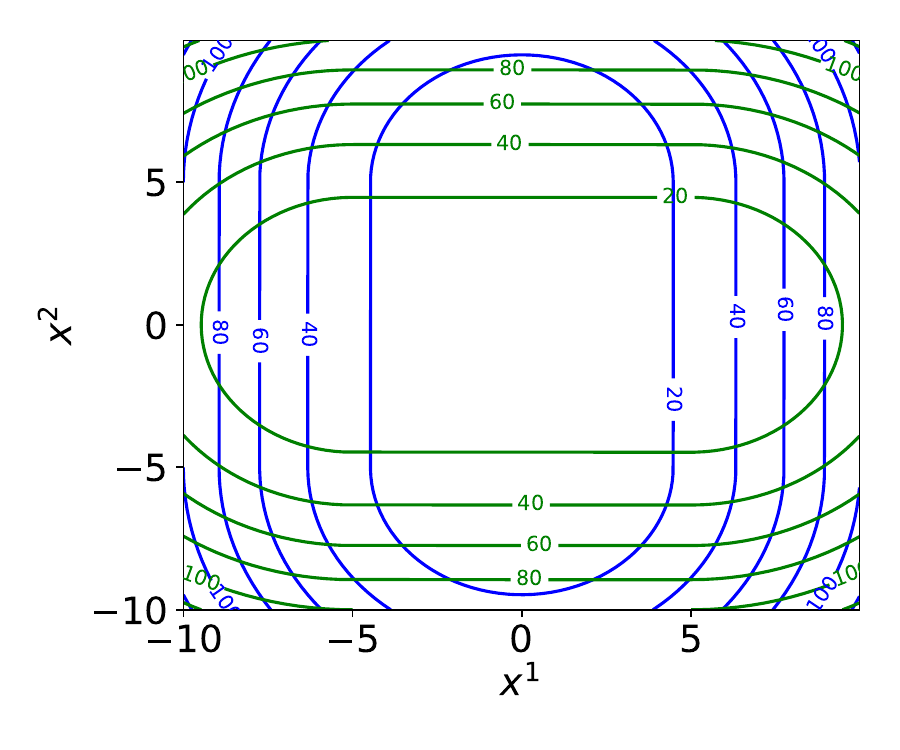}
\caption{Reward landscape in 2D space.}
\label{fig:mesh1}
\end{wrapfigure}

\begin{figure}[!htb]
   \begin{subfigure}{0.332\textwidth}
     \centering
     \includegraphics[width=.99\linewidth]{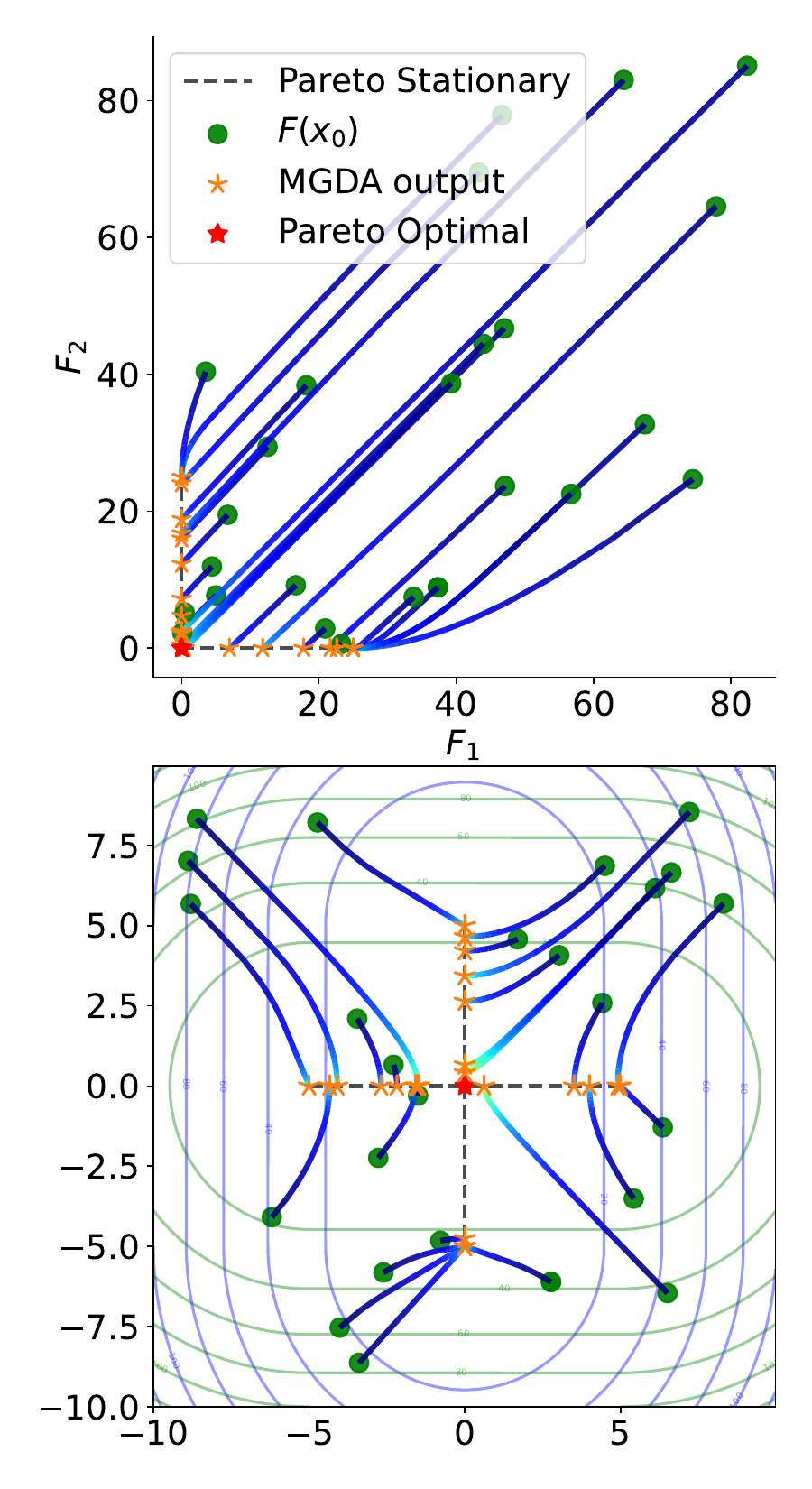}
     \caption{MGDA}\label{Fig:mgda}
   \end{subfigure}\hfill
   \begin{subfigure}{0.332\textwidth}
     \centering
     \includegraphics[width=.99\linewidth]{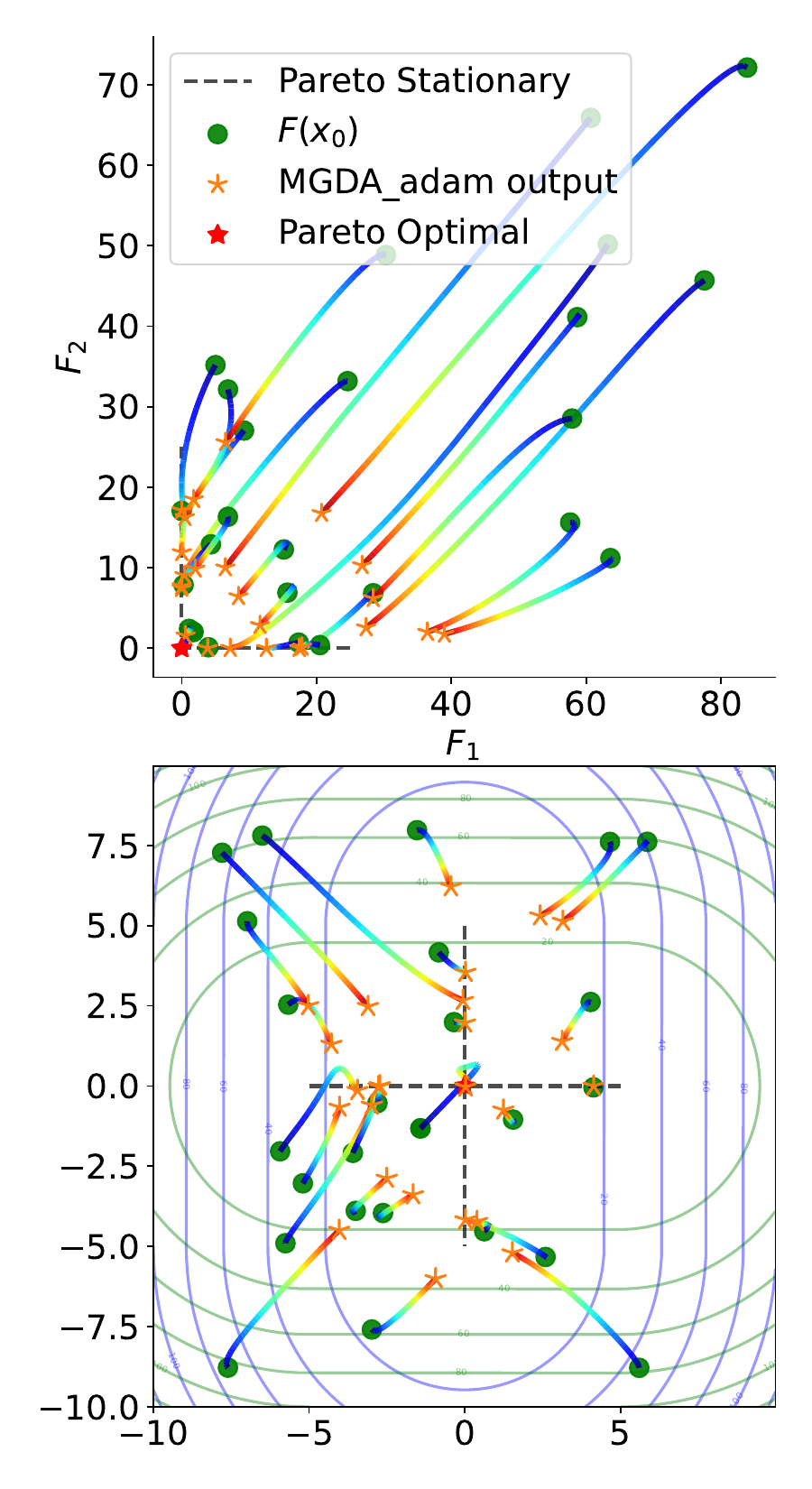}
     \caption{MGDA w/ Adam}\label{Fig:mgda++_adam}
   \end{subfigure}\hfill
   \begin{subfigure}{0.332\textwidth}
     \centering
     \includegraphics[width=.99\linewidth]{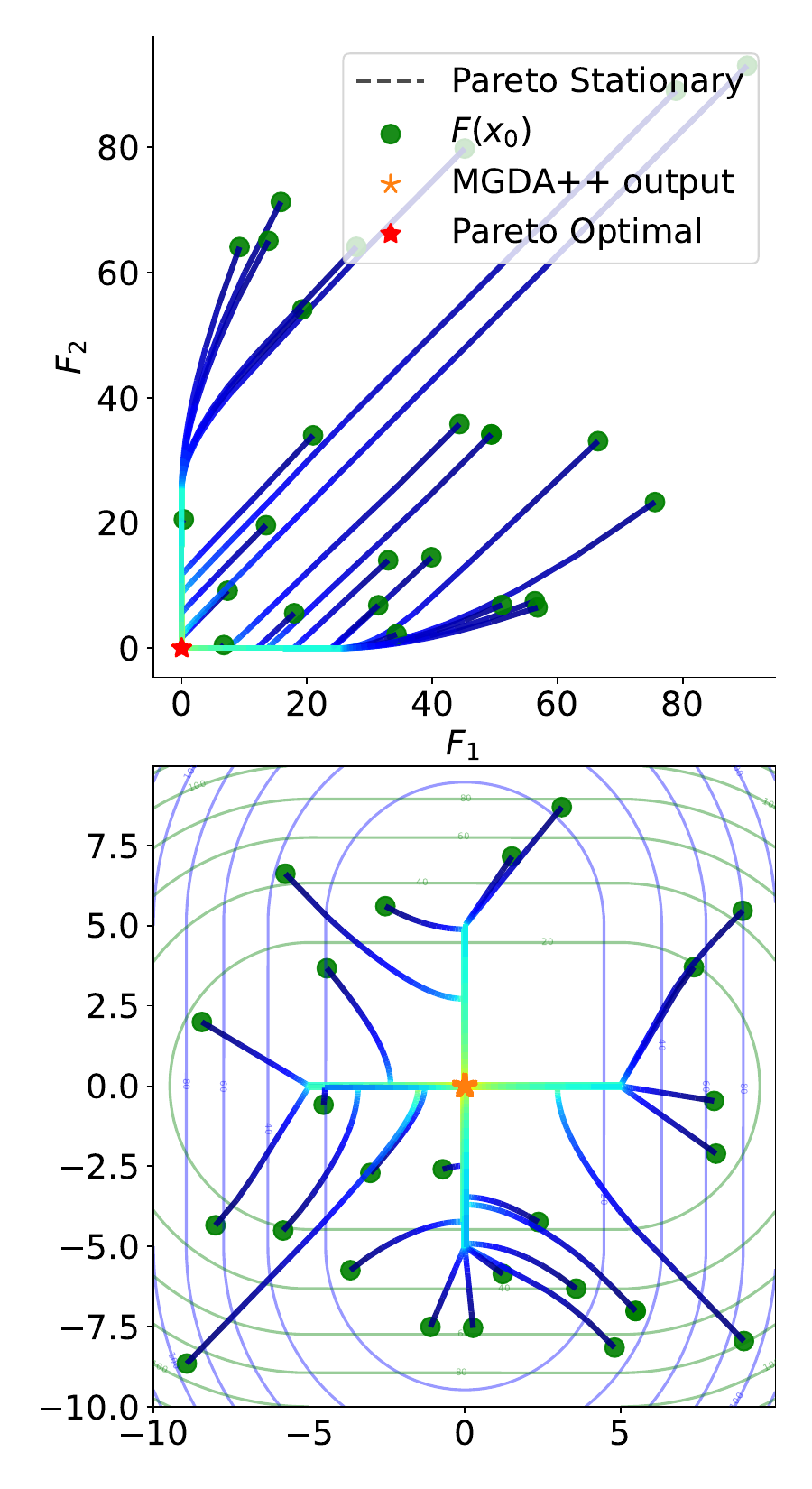}
     \caption{MGDA++}\label{Fig:mgda++}
   \end{subfigure}
   \caption{Comparison of MGDA, MGDA with Adam and MGDA++. Left: MGDA gets stuck at Pareto Stationary points, where the learning completely stops. Middle: Changing the optimizer does not help in avoiding the suboptimal convergence. Right: MGDA++ is able to converge to strong Pareto Optimal solutions while avoiding being trapped at Pareto Stationary points.} 
   \label{fig:compare_mgda_mgdapp}
\end{figure}

It is known that MGDA can converge to Pareto stationary solutions \citep{desideri2012multiple}. However, as illustrated in Figure \ref{fig:compare_mgda_mgdapp}, a stationary point can be a weak Pareto Optimal point and thus can be very suboptimal compared to the strong Pareto one. This is a distinct difference from the single objective optimization; in the convex case, stationary points are global optimal solutions, whereas in MOO, stationary points only ensure weak optimality, as demonstrated by the following lemma \citep{zeng2019convergence}.
\begin{lemma}[{Theorem 3.3, \cite{zeng2019convergence}}]
    When the objective functions are convex, any Pareto stationary points $\hat x$ are weak Pareto optimal solutions.
\end{lemma}
As a result, MGDA is only able to converge to weak Pareto optimal solutions \citep{fliege2019complexity}. 
Due to the optimization of the search directions in MGDA, the algorithm stops at any weak Pareto points. This stopping criterion can cause the algorithm to terminate prematurely or worse, not learn at all just by adding additional spurious objectives. We illustrate this phenomenon through the following corollary.
\begin{corollary}
    Given any optimization problem with the objective function $F:\mathbb R^m\rightarrow \mathbb R^n$, by adding a dummy objective $F_{n+1}(x)=c, \forall x$ with $c$ a constant number, then all the points $x$ in the new problem are weakly Optimal. As a result, the MGDA algorithm is not working in the modified problem.
\end{corollary}
We further show the weak optimal convergence of MGDA through a toy experiment in Figure \ref{fig:compare_mgda_mgdapp} with the bi-objective convex landscape in Figure \ref{fig:mesh1}, defined as
\[F_i(x) = \|y\|^2, \quad\text{with } y^j = \begin{cases}
    x^j &\text{if } j \neq i \\
    \max(0, |x^j| - 5) &\text{if } j = i
\end{cases}\]

The weak Pareto optimal convergence of MGDA can be attributed to the heavy influence of the diminishing-norm gradients from the already converged objectives to the common descent direction.
To elaborate, as gradients approach zero, the weights vector in the sum of the problem (\ref{linp:mgda}) becomes highly concentrated on the corresponding entries. While the impact of small gradients is also noted in \cite{desideri2012multiple}, attributing to the slow convergence rate of practical MGDA. 
We further associate it with the weak Pareto convergence problem. 
Furthermore, while the recommended approach to normalize the gradient norms all to be equal can somewhat mitigate the weak convergence issue of MGDA indirectly, 
its efficacy remains largely empirical, lacking solid theoretical grounding within MGDA's framework.
Rather, it stands as an implementation detail for improving convergence speed in practice.

\subsection{MGDA++ Algorithm}

In Lemma \ref{lem:sufficientPareto}, we establish a crucial criterion for strong Pareto optimality when the convex sum of \textit{non-zero} gradients equals zero. 
This insight suggests that a specific rooting of the problem of MGDA might be due to the diminishing gradient norm of some specific objectives that impede the optimization of others, and hints at the potential enhancement to MGDA by carefully treating the objectives that already converge. This motivates us to propose an improved version of MGDA, which removes the effect of small gradient norms on the optimization of other non-converging objectives by considering in (\ref{linp:mgda}) only a subset of gradients whose norm is greater than a certain threshold. The new algorithm, denoted as MGDA++, is described in the Algorithm \ref{alg:main}.



\begin{algorithm}
\caption{MGDA++ Algorithm}\label{alg:main}
\hspace*{\algorithmicindent} \textbf{Input}: $\epsilon > 0$, initial solution $x_0$
\begin{algorithmic}[1]
\For{$k=0, 1, \dots$}
    \State $S_k \leftarrow \emptyset$
    \For{$i=1, \dots, n$}
        \State Calculate $\nabla F_i(x_k)$
        \If{$\|\nabla F_i(x_k)\|>\epsilon$}
            \State $S_k = S_k \cup \{i\}$
        \EndIf
    \EndFor
    \If{$S_k=\emptyset$}
        \State Stop
    \EndIf
    \State Find $\{\lambda_i\}_{i\in S_k}$ by solving (\ref{linp:mgda}) on the subset of gradients $\{\nabla F_i(x_k)\}_{i\in S_k}$
    \State $d_k\leftarrow \sum_{i\in S_k}\lambda_i \nabla F_i(x_k)$
    \State Choose step size $t_k$
    \State $x_{k+1}\leftarrow x_k - t_k d_k$
\EndFor
\end{algorithmic}
\end{algorithm}


Below, we provide the theoretical analysis of the convergence property of our proposed MGDA++. Within our analysis, we make the following assumptions,
\begin{assumption}\label{ast:cvx_smt}
    All objective functions $F_i$ are convex and $L$-smooth.
\end{assumption}
\begin{assumption}\label{ast:opt}
    With a given point $x_0$, the sum of objectives function $F$ has a bounded level set $\Gamma=\{x|\sum_i F_i(x) \leq \sum_iF_i(x_0)\}$.
    Furthermore, 
    the optimal set of each objective function $F_i$ is non-empty and is denoted by $X_i^*$,
    the optimal value is denoted similarly as $F_i^*$.
\end{assumption}
First, we introduce a sufficient condition for a solution to be Pareto optimal when the objective functions are convex.
It is important to distinguish this condition from the widely acknowledged necessary conditions for Pareto optimality \citep{desideri2012multiple}. 
{Note that this and the next proposition hold true for a general number of tasks.}
\begin{lemma}\label{lem:sufficientPareto}
    Under assumption \ref{ast:cvx_smt}, if there exists a convex combination of the subset of {all non-zero} gradient vectors
    \begin{equation} \label{eq:pareto_contidion}
        \sum_{i\in S} \lambda_i \nabla F_i(x)=0 ; \quad \lambda_i > 0, \quad \|\nabla F_i(x)\| > 0 \quad \forall i \in S
    \end{equation}
    with $S \subseteq [n], S \neq \emptyset$, then $x$ is Pareto optimal.
\end{lemma}
\begin{proof}
    We only need to prove that $x$ is Pareto optimal on the subset of the $S$ objectives, {since objectives with zero gradients are already optimal}. 
    By contradiction, assume that $x$ is a point that satisfies condition \ref{eq:pareto_contidion} and there exist some $x^*$ such that $F_S(x^*)\preceq F_S(x)$ and $F_S(x^*) \neq F_S(x)$. We will show that all the gradient vectors of the objectives in $S$ lie in one half-plane whose orthogonal vector is $x^*-x$.

    Let $v = x^* - x$, then $v\neq 0$. Since all function $F_i$ are convex, then,
    \[F_i(x + tv) \leq (1-t)F_i(x) + t F_i(x^*)\]
    with some $t\in (0, 1]$. Divide both sides with $t$ and take the limit toward 0, we have
    \[\lim_{t\rightarrow 0}\frac{F_i(x + tv)-F_i(x)}{t}\leq F_i(x^*) - F_i(x) \leq 0, \quad \forall i \in S\]
    All the directional derivatives of $F_i$ along the vector $v$ are non-positive, which means that they all belong to the same half-space perpendicular to the vector $v$. Furthermore, since $F_S(x^*) \neq F_S(x)$, some gradient vectors are strictly inside the half-space.

    It is easy to see that any convex sum of non-zero vectors in a half-space, with some strictly inside, cannot equal zero, so this contradicts the condition in \ref{eq:pareto_contidion}. As a result, no such $x^*$ exists, and thus $x$ is Pareto optimal by definition.
\end{proof}

Next, we show that an arbitrary approximation of Pareto optimal solutions can be obtained if the gradients of all objectives are small enough, 

\begin{proposition} \label{prop:small_eps}
    Under assumptions \ref{ast:cvx_smt} and \ref{ast:opt}, then for any $\varepsilon>0$, it is possible to choose an $0 < \epsilon \leq \sqrt{2L\varepsilon}$ such that if any $x$ satisfies $\|\nabla F_i(x)\|<\epsilon$, then $F_i(x) \leq F_i^* + \varepsilon, \forall i \in [n]$. Such $x$ are necessarily $\varepsilon$-Pareto optimal solutions.
\end{proposition}
Before presenting the proof, we first remind two basic properties of convex, smooth function. For a convex, $L$-smooth function $f$ and two $x, y \in \text{dom}(f)$, then the followings hold true
\begin{equation}\label{eq:prop1}
    f(y) \leq f(x) + \langle \nabla f(x), y-x\rangle + \frac{L}{2}\|x-y\|^2
\end{equation}
\begin{equation}\label{eq:prop2}
    f(y) \geq f(x) + \langle \nabla f(x), y-x\rangle + \frac{1}{2L}\|\nabla f(x) - \nabla f(y)\|^2
\end{equation}
These properties are well known and the derivation can be found in optimization textbooks (e.g. Theorem 5.8 \citep{beck2017first}). 

\begin{proof}[Proof of Proposition~{\upshape\ref{prop:small_eps}}]
    From (\ref{eq:prop2}), for each $i\in [n]$, let $x_i^{*} \in X^*$ and plug $x_i^{*}$ and $x$ into $y$ and $x$ respectively
    \[F_i(x_i^*) \geq F_i(x) + \langle \nabla F_i(x), x^*_i-x\rangle + \frac{1}{2L}\|\nabla F_i(x)\|^2\]
    In order for $F_i(x^*_i) \geq F_i(x) - \varepsilon$ for any $x$, define $B=\|x^*_i-x\|$, we need that
    \begin{align*}
        &\langle \nabla F_i(x), x^*_i-x\rangle + \frac{1}{2L}\|\nabla F_i(x)\|^2 \\
        \geq & \frac{1}{2L}\|\nabla F_i(x)\|^2 - \|\nabla F_i(x)\|\|x^*_i-x\|  \tag{Hölder's inequality} \\
        = & \frac{1}{2L}\|\nabla F_i(x)\|^2 - B\|\nabla F_i(x)\| 
        \geq -\varepsilon
    \end{align*}
    which is a quadratic function of $\|\nabla F_i(x)\|$. If $\varepsilon \geq B^2L/2$, then it is satisfied with any value of $\|\nabla F_i(x)\|$,
 solving this with $\varepsilon \leq B^2L/2$ yields
    \begin{numcases}{}
        \|\nabla F_i(x)\| \geq L\left(B+\sqrt{B^2 - \frac{2}{L}\varepsilon}\right)\label{eq:case1}\\
        \|\nabla F_i(x)\| \leq L\left(B-\sqrt{B^2-\frac{2}{L}\varepsilon}\right)\leq \sqrt{2L\varepsilon}\label{eq:case2}
    \end{numcases}
    we then show that (\ref{eq:case1}) is an invalid choice of $\|\nabla F_i\|$ and would never happen under assumption \ref{ast:cvx_smt}.

    Plug $x$ and $x^*_i$ into $y$ and $x$ in both (\ref{eq:prop2}) and (\ref{eq:prop1}), we get
    \[\frac{1}{2L}\|\nabla F_i(x)\|^2 \leq F_i(x) - F_i(x^*_i)\leq \frac{L}{2}\|x-x^*_i\|^2\]
    \[\Rightarrow \|\nabla F_i(x)\| \leq L\|x-x^*_i\|= LB\]
    this invalidates the choice (\ref{eq:case1}), then we are left with the case (\ref{eq:case2}). As a result, by choosing $\epsilon\leq\sqrt{2L\varepsilon}$, then any $x$ with $\|\nabla F_i(x)\|\leq \epsilon$ will satisfy that $F_i^* \geq F_i(x)-\varepsilon$, this concludes the proof.
\end{proof}
Now we are ready to present our main theoretical result of the convergence of MGDA++ with two tasks.

\begin{figure}[!htb]
   \begin{minipage}{0.499\textwidth}
     \centering
     \includegraphics[width=.99\linewidth]{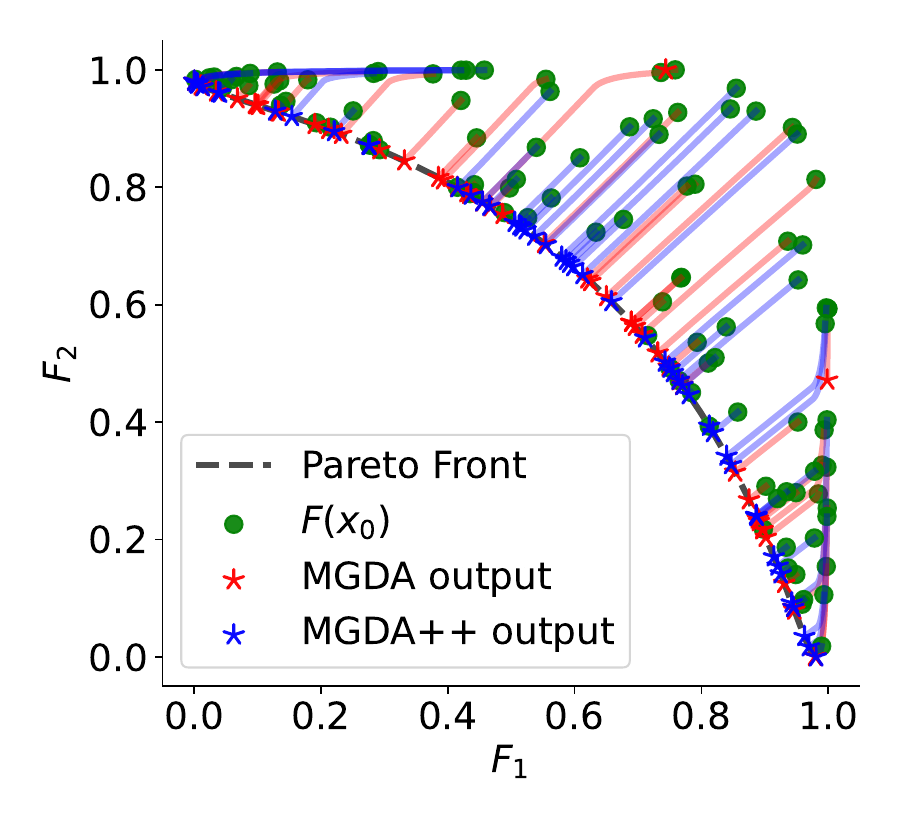}
   \end{minipage}\hfill
   \begin{minipage}{0.499\textwidth}
     \centering
     \includegraphics[width=.99\linewidth]{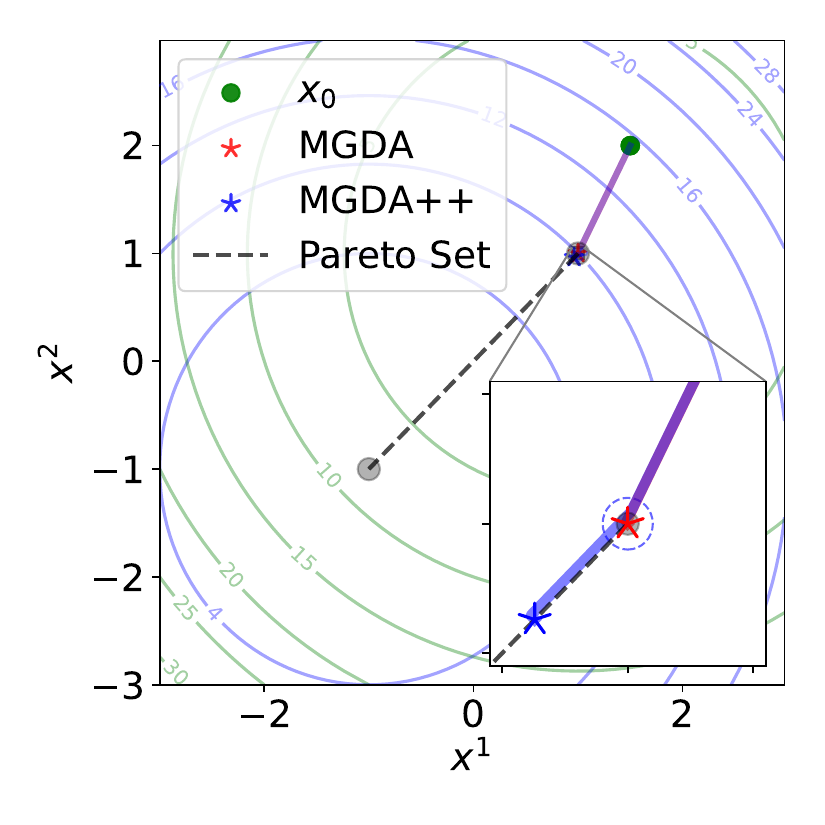}
   \end{minipage}
   \caption{Comparison of MGDA and MGDA++ convergent points. Left: We test MGDA and MGDA++ on the synthetic problem from \cite{lin2019pareto}. Both algorithms can converge to different Pareto optimal solutions in most usual cases. Right: We initiate the two algorithms with the same starting point in a simple quadratic bi-objective optimization problem, $F_{1, 2}(x)=\|x\pm \mathbf{1}\|^2$. While MGDA stops as soon as it reaches the first stationary point, MGDA++ avoids small gradient norm solutions by further taking an additional step into the relative interior of the Pareto Set. In this example, MGDA++ does not converge to balls with radius $\epsilon/2$ around stationary points whose gradient norms of one of the objectives equal 0. For visualization, we plot such a ball with doubled radius $\epsilon=0.01$ as a blue-dotted empty circle in the figure.}
   \label{fig:compare_mgda_mgdapp_convergence}
\end{figure}

\begin{theorem}\label{theo:convergence}
    Under assumptions \ref{ast:cvx_smt} and \ref{ast:opt}, for any $\varepsilon > 0$, with $n=2$ and by choosing $\epsilon < \sqrt{2L\varepsilon}$ and with appropriate choices of each update steps $t_k$ as
    \begin{equation*}
         t_k = \begin{cases}
         \displaystyle\max\left(\frac{|S_k|\|d_k\|^2 + \big\langle \sum_{i \in \overline{S}_k}\nabla F_i(x_k), d_k\big\rangle}{nL\|d_k\|^2}, 0\right) & \text{if } \|d_k\| > 0\\
             0 & \text{if } \|d_k\|=0
         \end{cases},
     \end{equation*}
    with $\overline{S}_k$ the complement of $S_k$, then each convergent subsequence of MGDA++ converges to either Pareto optimal or $\varepsilon$-Pareto optimal solutions.
\end{theorem}
\begin{proof}[Proof of Theorem~{\upshape\ref{theo:convergence}}]
     At the iteration $k$, let $S_k$ the set of indices for which $\|\nabla F_s(x_k)\| > \epsilon, \forall s \in S_k$ and $\overline{S}_k$ its complement ($|S_k|\leq 2$). We also assume that the set $S_k$ is non-empty, or else all the gradient vectors have norms less than $\epsilon$ and the algorithm would terminate. Then, let $d_k$ the solution to the problem (\ref{eq:mgda_prob}) with the set of $S_k$ objectives, we have the following observations  
     \[F_i(x_{k+1}) \leq F_i(x_k)+\langle \nabla F_i(x_k), x_{k+1}-x_k \rangle +\frac{L}{2}\|x_k-x_{k+1}\|^2, \quad i \in \overline{S}_k\]
     \[F_j(x_{k+1}) \leq F_j(x_k) -t_k\|d_k\|^2+\frac{L}{2}t_k^2\|d_k\|^2, \quad j \in S_k\]
     with $x_{k+1}-x_k=-t_k d_k$, summing over all indices, we get
     \[(F_1+F_2)(x_{k+1})\leq (F_1+F_2)(x_{k}) + {L}t_k^2\|d_k\|^2 - t_k|S_k|\|d_k\|^2 - t_k\bigg\langle \sum_{i \in \overline{S}_k}\nabla F_i(x_k), d_k\bigg\rangle\]
     As a result, if we choose 
     \begin{equation*}
         t_k = \begin{cases}
             \max\left(\frac{|S_k|\|d_k\|^2 + \big\langle \sum_{i \in \overline{S}_k}\nabla F_i(x_k), d_k\big\rangle}{2L\|d_k\|^2}, 0\right) & \text{if } \|d_k\| > 0\\
             0 & \text{if } \|d_k\|=0
         \end{cases}
     \end{equation*}
     then the sum of all the objectives is guaranteed to non-increase after each iteration. Let $f_k = (F_1+F_2)(x_{k})$, $f_k \geq \sum_i F^*_i$, the sequence $f_0, f_1, \dots$ is monotonically non-increasing and bounded below, so it converges to some $f^*$, since $\{x_k\}$ is bounded by assumption \ref{ast:opt}, there exists a subsequence that converges to some $\widetilde{x}$, 

     \begin{enumerate}
         \item If at $\widetilde{x}$, all the gradient vector norms are less than $\epsilon$, then the algorithm terminates at $\widetilde x$, and by choosing small enough $\epsilon$, by proposition \ref{prop:small_eps}, the algorithm converges to arbitrary $\varepsilon$-Pareto optimal solutions.

         \item Otherwise, let $\widetilde S$ the set of indices for which $\|\nabla F_i(\widetilde{x})\|>\epsilon, \forall i \in \widetilde S$, then $\widetilde S$ is nonempty, we also define $\widetilde d$ as the solution of the problem \ref{eq:mgda_prob} at $\widetilde x$. We consider two possible cases.
         
         \begin{itemize}
         \item [a)] If $\|\widetilde d\| = 0$, then since $\widetilde d$ is the optimal solution of \ref{eq:mgda_prob}, then $0=\widetilde d=\sum_{i\in S^*} c_i \nabla F_i(\widetilde x), \sum_i c_i=1, c_i\geq 0$ on the subset $\widetilde S$ of non-zero gradient vectors, and according to Lemma \ref{lem:sufficientPareto}, $\widetilde x$ is Pareto optimal.
         \item [b)] If $\|\widetilde d\| > 0$, then the corresponding update step $\widetilde t = 0$ or else by updating $x' = \widetilde{x} - \widetilde{t} \widetilde{d}$, then $(F_1 + F_2)(x') < f^*$. Since $\widetilde t = 0$, we have that
         \[|\widetilde S|\|\widetilde d\|^2 + \big\langle \sum_{i \in \overline{\widetilde S}}\nabla F_i(\widetilde x), \widetilde d\big\rangle \leq 0\]
         and because $\|\widetilde d\| > 0$, so $|\widetilde S|=1$. Let $j$ be the only index in $\overline{\widetilde S}$, then
         \begin{align}
             &\|\widetilde d\|^2 + \langle\nabla F_j(\widetilde x), \widetilde d \rangle \leq 0 \notag \\
             \Rightarrow & \|\widetilde d\|^2 - \|\nabla F_j(\widetilde x)\| \|\widetilde d\| \leq 0 \notag \\
             \Rightarrow & \|\widetilde d\| \leq \|\nabla F_j(\widetilde x)\|. \label{eq:conv2b}
         \end{align}
         However, since $|\widetilde S|=1$, let $k$ the only index in $\widetilde S$, then $\widetilde d = \nabla F_k(\widetilde x)$. By definition of $\widetilde S$, $\|\widetilde d\|=\|\nabla F_k(\widetilde x)\| > \epsilon$, but $\|\nabla F_j(\widetilde x)\| \leq \epsilon$, which contradicts with (\ref{eq:conv2b}). As a result, this case cannot happen.
         \end{itemize}
     \end{enumerate}
In all the possible cases, we have shown that the algorithm converges to either Pareto or $\varepsilon$-Pareto optimal solutions, this concludes the proof.
\end{proof}
\begin{remark}[Update step sizes]
    At some iteration, if $\overline{S}_k = \emptyset$ (i.e. no gradient vectors are dropped), then the choice of the update steps $t_k=\frac{1}{L}$, which is the optimal update step for convex, smooth functions. In this case, MGDA++ returns to the normal MGDA with a constant step size. Empirically, we find that MGDA++ works fine with constant step sizes in practice.
\end{remark}
\begin{remark}[Challenges when generalizing to general $n>2$] The convergence analysis of MGDA++ is hard for several reasons: first, each objective is not monotonically decreasing, because when the gradients become small, they are removed from consideration of steepest decent direction. Secondly, gradient norms do not monotonically decrease with arbitrary update step sizes. Finally, the algorithm does not converge to a single optimal solution $x^*$ nor objective vector $F^*$. 
In the proof of Theorem \ref{theo:convergence}, we specifically 
rely on the monotonicity of the sum of all objective functions by carefully selecting step sizes. 
    The proof generally fails to generalize to other numbers of $n$, as when $|S_k|>1$, the update vector $d$ can have a small norm $< \epsilon$ even when all other gradient vectors in $S_k$ have norm $>\epsilon$, which cannot ensure the monotonicity of $\{f\}_k$. 
\end{remark}
\begin{remark}[Convergent solutions]
    MGDA++ guarantees strong optimality by avoiding potentially suboptimal solutions with small gradient norms, illustrated as the case 2b) in the proof of theorem \ref{theo:convergence}. 
    However, some of these potential solutions are actually Pareto optimal. As a result, MGDA++ only recovers a subset of the Pareto set where the optimality is ascertained. We demonstrate this phenomenon in Figure (\ref{fig:compare_mgda_mgdapp_convergence}). We believe this is inevitable without any look-ahead search while keeping the optimal convergence guarantee.
\end{remark}

Theorem \ref{theo:convergence} establishes strong Pareto optimal convergence with convex objective functions. 
To the best of our knowledge, even within the limited setting of bi-objective optimization, this is the first strong Pareto convergence instance of gradient descent methods with convex functions in the literature.

\section{Experiments}

\begin{figure}[!htb]
   \begin{subfigure}{.48\linewidth}
    \centering
    \includegraphics[width=.99\linewidth]{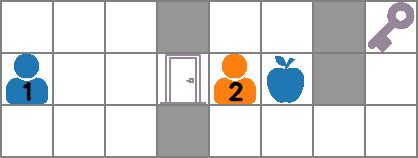}
    \caption{Door}
\end{subfigure}\hfill
   \begin{subfigure}{.48\linewidth}
    \centering
    \includegraphics[width=.99\linewidth]{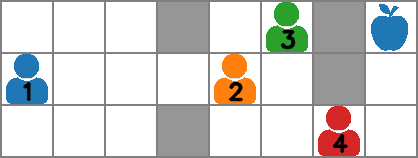}
    \caption{Dead End}
\end{subfigure}\hfill
\begin{subfigure}{.48\linewidth}
    \centering
    \includegraphics[width=.99\linewidth]{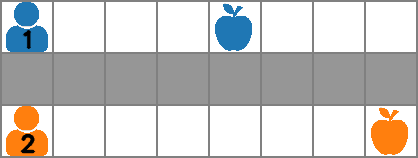}
    \caption{Two Corridors}
\end{subfigure}\hfill
\begin{subfigure}{.48\linewidth}
    \centering
    \includegraphics[width=.99\linewidth]{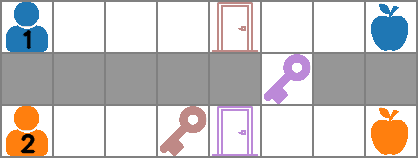}
    \caption{Two Rooms}
\end{subfigure}\hfill
   \caption{Four scenarios of the Gridworld environment.}
   \label{fig:gridworld}
\end{figure}

\textbf{Gridworld}. 
We consider the grid world environments, inspired by the altruistic learning environments setup \citep{franzmeyer2022learning}, where each agent can navigate in a grid with four directions. 
Agents receive a big reward ($+10$) when reaching the goal position (depicted as an apple) with matching colors for the first time in each episode. 
Conversely, collisions with walls or other agents result in a small penalty ($-0.1$).
Furthermore, there are doors that agents cannot pass through, which can be opened as long as some agent occupies the key position with the same color as the doors. 
We investigate four variations of the environment {(Figure }\ref{fig:gridworld}{)}:
\begin{enumerate}
    \item [a)] \textbf{Door} \citep{franzmeyer2022learning}. This environment requires the second agent to open the door for the first one so that it can reach its goal. However, this action does not gain any benefit for that agent. In this environment, selfish learners are expected to fail, while other altruistic learning paradigms will succeed. This environment illustrates the advantage of altruistic behaviors.
    \item [b)] \textbf{Dead End} \citep{franzmeyer2022learning}. In this scenario, the first agent must navigate to the goal located at the far end of the map, but there are several agents that can block the path of the first agent. This problem illustrates the optimization landscape of MARL that can be complex and dynamical due to the interactions and exploration of other agents. This is a distinction from other settings, where one agent's learning can interfere with the others' optimization problems.
    \item [c)] \textbf{Two Corridors}. Two agents are located at separate corridors, each with their respective goals located on the other side. This scenario does not require any cooperation between agents. 
    However, their tasks vary in difficulty.
    While the first agent traverses only half the corridor, the second agent explores to the full end. 
    The asymmetric level of exploration means one agent will converge faster than the other one. 
    Once the easier task agent converges, every policy from the harder task agent becomes weak Pareto optimal. 
    This example illustrates the prevalence of weakly optimal solutions in MARL with diverging convergent rates, particularly in tasks with varying difficulty levels. 
    \item [d)] \textbf{Two Rooms}. This environment studies the implicit cooperation. Two agents are trapped in two different rooms. While both agents cannot open their doors themselves, the first one can open the door of the second. 
    Upon escaping, the second agent can choose to reciprocate by opening the first agent's door or pursuing its own objectives, potentially betraying its benefactor. 
    However, opening the second agent's door is always optimal from the first agent's perspective, as it is the only way to mutual escape. 
    This optimal behavior is not explicitly incentivized in the reward scheme, as helping the other does not directly contribute to finding the goal.
\end{enumerate}

\begin{center}
    \begin{table}[!htb]
        \begin{tabular}{ll|ccccc}
    \toprule 
    Scenario& agent& MGPO++ & MGPO & MAPPO & IQL & IPPO \\\midrule
    \multirow{2}{4em}{Door} &1 & $\mathbf{10.0\pm0.0}$ & $\underline{9.9 \pm 0.0}$ & $0.0\pm0.0$ & $-0.0 \pm0.0$ & $-0.0 \pm 0.0$\\
     &2& $\mathbf{-0.0\pm0.0}$ & $-6.9\pm4.9$ & $\mathbf{-0.0\pm0.0}$ & $\underline{-0.1\pm0.0}$ & $\mathbf{-0.0\pm0.0}$ \\\midrule
     \multirow{4}{4em}{Dead End}&1 & $\mathbf{3.3\pm4.7}$ & $0.0\pm0.0$ & $0.0\pm0.0$ & $\underline{0.3\pm0.4}$ & $0.0\pm0.0$ \\
     &2& $\mathbf{-0.0\pm0.0}$ & $-14.4\pm0.8$ & $\underline{0.0\pm0.0}$ & $-0.0\pm0.0$ & $\underline{0.0\pm0.0}$ \\
     &3& $\mathbf{-0.0\pm0.0}$ & $-13.9\pm2.2$ & $\mathbf{-0.0\pm0.0}$ & $\underline{-0.1\pm0.0}$ & $\mathbf{-0.0\pm0.0}$ \\
     &4& $\underline{-0.0\pm0.0}$ & $-14.7\pm2.2$ & $\mathbf{0.0\pm0.0}$ & ${-0.1\pm0.0}$ & $\underline{-0.0\pm0.0}$\\ \midrule
     \multirow{2}{4em}{Two Corridors}&1 & $\mathbf{10.0\pm0.0}$ & $\mathbf{10.0\pm0.0}$ & $\mathbf{10.0\pm0.0}$ & $\underline{9.9\pm0.0}$ & $\underline{9.9\pm0.0}$\\
     &2& $\mathbf{10.0\pm0.0}$ & $-1.1\pm1.6$ & $\mathbf{10.0\pm0.0}$ & $9.9\pm0.0$ & $9.9\pm0.1$\\\midrule
     \multirow{2}{4em}{Two Rooms}&1& $\mathbf{10.0\pm0.0}$ & $\mathbf{10.0\pm0.0}$ & $-0.0\pm0.0$ & $-5.6\pm8.0$ & $\underline{6.5\pm3.6}$ \\
     &2& $\mathbf{9.8\pm0.2}$ & $-17.7\pm0.9$ & $\underline{0.0\pm0.0}$ & $-0.1\pm0.0$ & $-0.0\pm0.0$ \\\botrule
    \end{tabular}
        \caption{Result on Gridworld environment, all reported values are rounded to the first place after decimal. The best values of each agent are highlighted in bold, while the second-best values are underscored.}
        \label{tab:gridworld}
    \end{table}
\end{center}


We compare our method with the following baselines. 
\begin{enumerate}
    \item \textbf{Multi-head MAPPO}. We extend MAPPO \citep{yu2022surprising} to work in cooperative environments with different reward functions for each agent. As described in section \ref{sec:3.1}, the algorithm is similar to MAPPO, the differences are that the output of the value function is multi-headed, and each head corresponds to predicting the expected returns of the respective agents. The value network is shared between agents and can access global information during training, adhering to the Centralized training decentralized execution (CTDE) paradigm. 
    \item \textbf{IPPO} \citep{de2020independent} decomposes the learning of all agents in a completely decentralized manner, where each of them is a single, independent PPO learner. The learning of IPPO disregards the dynamic of all other agents, attributing them to the non-stationarity of the environment. However, with appropriate setups, IPPO can perform competitively with other centralized training methods \citep{sun2022trust}.
    \item \textbf{IQL} \citep{tampuu2017multiagent} is the extension of the DQN architecture to multiagent environments. Similar to IPPO, IQL also learns decentrally with independent Q networks and separate replay buffers for each agent. Compared to other methods, IQL is the only off-policy, value-based method that we consider in our work.
    \item \textbf{\underline{MG}DA w/ MAP\underline{PO} (MGPO)}. We use the multi-head architecture of MAPPO, but instead of individually optimizing the objective of each agent separately, we use MGDA to find a common direction for all the agents as discussed in section \ref{sec:3.1}. 
    \item \textbf{\underline{MG}DA++ w/ MAP\underline{PO} (MGPO++)}. This setting is similar to MGPO, but we use our proposed MGDA++ method in place of MGDA. 
\end{enumerate}

\textbf{Experimental details}. 
For consistency, all methods utilize three-layer MLPs with hidden sizes of 64. Hyperparameters of MAPPO, IPPO, and IQL follow the default settings of their repositories. 
MGPO and MGPO++ adopt MAPPO's hyperparameters.
In MGDA++, $\epsilon$ is set to $0.05$, except in Dead End where it is $0.1$.
In each scenario, we train all the methods for 500k environment steps. 
To ensure fair comparisons and a better analysis of the optimization methods, we avoid parameter sharing in the actor networks for on-policy methods and in the value networks for value-based methods. 
All experiments are repeated with three different seeds to improve evaluation reliability.

\begin{figure}[h]
    \centering
    \includegraphics[width=0.8\textwidth]{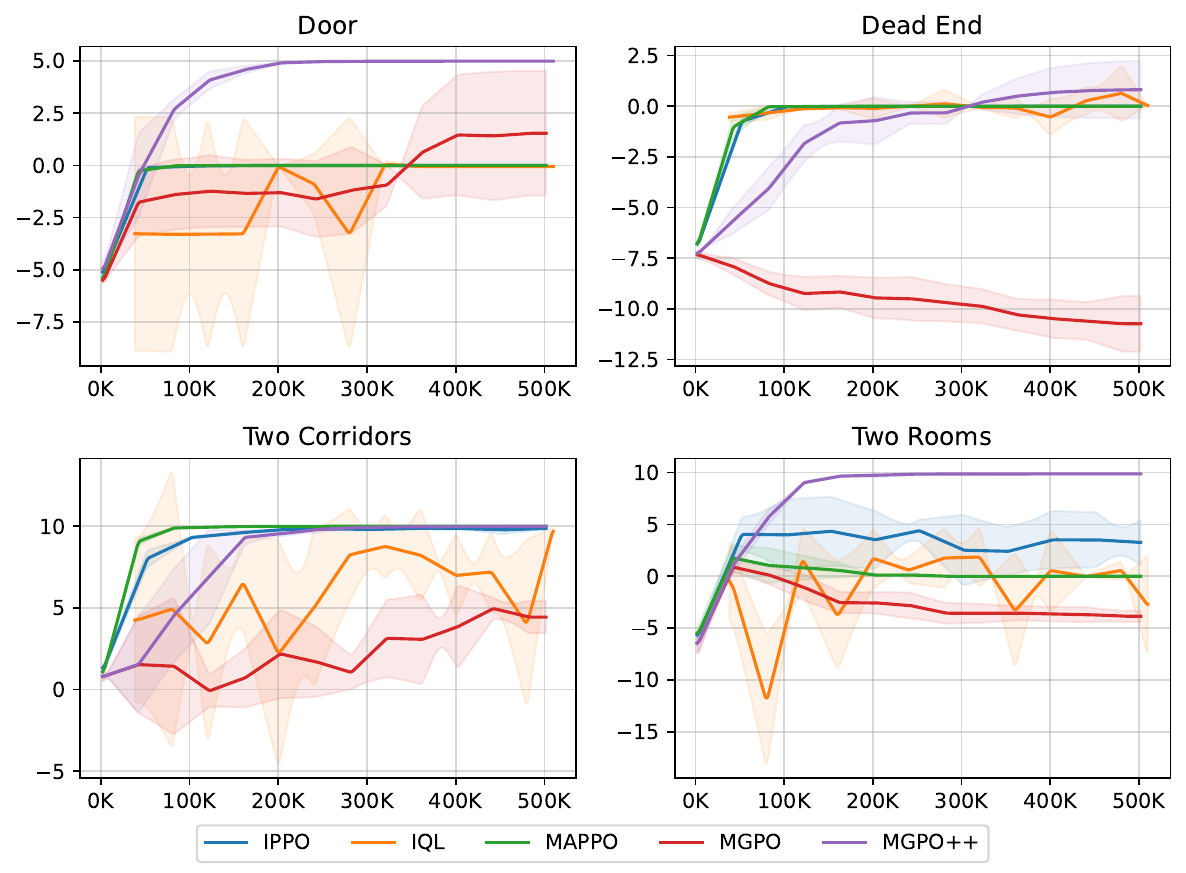}
    \caption{Average reward of all agents in the four scenarios. Note that, the averaging operator can be viewed as a particular linearization of the rewards vector.}
    \label{fig:res}
\end{figure}

\textbf{Experimental results.}
In Gridworld environments, we observe the superiority of our method to all other baselines (Table \ref{tab:gridworld} and Figure \ref{fig:res}). In the Door scenario, MAPPO with multi-objective optimization is the only method that can converge {to} the optimal solution for the first agent, while all other algorithms with single objectives do not learn to cooperate. In this case, we already see the advantage of MGDA++ compared to MGDA. While MGDA helps in finding the optimal policy for the first agent, the second agent is yet to converge at the time of convergence of the first one, after which this agent is unable to update due to the dominance of the diminishingly small gradient from the first agent, analyzed in the previous section. 

A similar result is observed in the Dead End scenario. Notably, MGPO does not learn except for the first agent, this highlights the fact that MGDA can only learn the optimal policy for only one agent. In MGPO++, we observe that all agents can learn good solutions without being hindered once some agents converge. We note that there are runs where the first agent is able to reach the goal, illustrating that the algorithm can find several local Pareto solutions, depending on the exploration of agents. This also demonstrates that MGDA++ can converge to strong Pareto solutions with a higher number of objectives ($n>2$). 

In the Two Corridors scenario, all the single objective methods converge as this scenario does not need cooperation. However, MGPO does not converge for the second agent, which is the harder task agent, since after the first agent converges, every policy from the second agent is weakly optimal. Hence, the algorithm stops after the first agent reaches optimal policy, even though the second agent is still learning. 

In the last scenario, only MGDA++ is able to reach the optimal policies for both agents. While MGDA can find high rewards for the first agent, the second agent is trapped after the first agent's convergence. The difference between MAPPO and IPPO is that we set the entropy coefficient of MAPPO to 0. While the first agent can find high rewards initially in MAPPO, this is only the artifact of the exploration of the second agent. As a result, when the entropy level of both agents decreases over time, the second agent's performance drops. On the other hand, the performance with IQL is unstable, possibly because of the non-stationarity perceived by the first agent due to the exploration effect of the second one. While this problem is also present in the on-policy methods, it is particularly persistent in the off-policy method because the collected data from the past in the replay buffer differs from the current policy. 

\section{Related works}

\textbf{Multi-agent Reinforcement learning}. 
While the concept of Pareto optimality is irrelevant in zero-sum games. 
In fully cooperative environments, 
there exist global, joint optimal policies that are better than any other policies as in the single agent setting \citep{sutton2018reinforcement}. The Pareto front in such cases thus comes down to contain a single optimal value function, leading to the equivalence of weak and strong Pareto solutions \citep{zhao2023local}. Dominant approaches in this setting include value decomposition \citep{rashid2020monotonic} and policy-gradient methods \citep{yu2022surprising}.

Early works in cooperative multi-agent settings involve training independent learners for each agent \citep{tan1993multi}. 
It was observed that independent learning could converge to Nash Equilibriums through Policy Gradient methods \citep{leonardos2021global}. 
With the actor-critic architecture, \cite{lowe2017multi} introduces centralized training to cooperative-competitive environments. 
More recently, \cite{zhao2023local} proposed a method for achieving global convergence to optimal policies in fully cooperative domains, and the highlighted suboptimality of current MARL methods by \cite{papoudakis2020benchmarking}. 
This has sparked interest in exploring Pareto optimal policies using multi-agent reinforcement learning techniques. 

To address this, \cite{christianos2022pareto} proposes Pareto actor-critic, which employs a modified advantage
estimation for directing the joint policy to Pareto equilibrium. However, this approach suffers from exponential computational costs and assumes a non-conflict nature of the agents' rewards.
Furthermore, it is restricted to single-reward domains, specifically fully cooperative environments.

\textbf{Multi-Objective/Multitask learning.}
While there are works that can find the strong $\varepsilon$-Pareto optimal solutions using gradient-free approaches \citep{evtushenko2014deterministic}, we focus on the gradient-descent based methods. 
\cite{desideri2012multiple} introduce the MGDA as steepest gradient descent \citep{fliege2000steepest} that converge to Pareto Stationary points. 
Subsequent future works include \cite{sener2018multi} that improves MGDA to be suitable for learning in high dimensional gradient in neural networks and \cite{lin2019pareto} for finding diverse Pareto solutions.

Regarding the optimality concept, many works do not distinguish strong and weak Pareto optimal solutions. For example, \cite{lin2019pareto, zhou2022convergence} define weak and strong Pareto optimality as similar concepts.
Other works assume additional conditions so that weak and strong Pareto optimal solutions coincide \citep{roy2023optimization}.
Pareto Stationary solutions are de facto in the theoretical results of MOO methods with gradient-based approaches \citep{fliege2019complexity, zeng2019convergence}.  

On the other hand, some other works explore the approaches of leveraging user preference directions to the Pareto optimality search \citep{momma2022multi}, such methods require additional knowledge of the preference vectors.
Linearization approaches are easier to optimize \citep{xiao2024direction}, but they are potentially incapable of fully exploring the Pareto front \citep{hu2024revisiting}. Other researches focus on finding the optimal solution on the average objectives while mitigating conflict between gradients \citep{yu2020gradient}, which does not account for a range of Pareto optimal solutions.
The proposed algorithm in this paper preserves the desired property of MGDA that can still explore most of the Pareto Front while improving on the strong Pareto convergence. 

\section{Conclusion}

In this work, we explore the possibility of training multi-agents in a cooperative scenario.
By viewing the vector-based reward as a multi-objective task, we can apply the well-known MGDA algorithm to the MARL domains.
However, the standard MGDA only guarantees to converge to a weak Pareto optimal solution.
Therefore, we introduce MGDA++, which filters out the small norm objective gradients to enhance convergence properties.
We also provide theoretical analysis for MGDA++ in bi-objective problems with convex and smooth settings.
The combination of MGDA++ and trust region MARL is evaluated in different scenarios in the Gridworld benchmark.
The results highlight that our proposed method can direct the agents' policy to strong Pareto optimality, outperforming various baselines.
In future works, one possible direction is to extend our theoretical analysis to a general number of objectives.
From the practical aspect, more complex scenarios and MARL benchmarks can be used to verify the effectiveness of the combination of MGDA++ and trust region framework based on the results of this paper.



\begin{appendices}






\end{appendices}

\bibliographystyle{plainnat}

\begin{thebibliography}{33}
\providecommand{\natexlab}[1]{#1}
\providecommand{\url}[1]{\texttt{#1}}
\expandafter\ifx\csname urlstyle\endcsname\relax
  \providecommand{\doi}[1]{doi: #1}\else
  \providecommand{\doi}{doi: \begingroup \urlstyle{rm}\Url}\fi

\bibitem[Beck(2017)]{beck2017first}
Amir Beck.
\newblock \emph{First-order methods in optimization}.
\newblock SIAM, 2017.

\bibitem[Cato(2023)]{cato2023weak}
Susumu Cato.
\newblock When is weak pareto equivalent to strong pareto?
\newblock \emph{Economics Letters}, 222:\penalty0 110953, 2023.

\bibitem[Christianos et~al.(2022)Christianos, Papoudakis, and Albrecht]{christianos2022pareto}
Filippos Christianos, Georgios Papoudakis, and Stefano~V Albrecht.
\newblock Pareto actor-critic for equilibrium selection in multi-agent reinforcement learning.
\newblock \emph{arXiv preprint arXiv:2209.14344}, 2022.

\bibitem[De~Witt et~al.(2020)De~Witt, Gupta, Makoviichuk, Makoviychuk, Torr, Sun, and Whiteson]{de2020independent}
Christian~Schroeder De~Witt, Tarun Gupta, Denys Makoviichuk, Viktor Makoviychuk, Philip~HS Torr, Mingfei Sun, and Shimon Whiteson.
\newblock Is independent learning all you need in the starcraft multi-agent challenge?
\newblock \emph{arXiv preprint arXiv:2011.09533}, 2020.

\bibitem[D{\'e}sid{\'e}ri(2012)]{desideri2012multiple}
Jean-Antoine D{\'e}sid{\'e}ri.
\newblock Multiple-gradient descent algorithm (mgda) for multiobjective optimization.
\newblock \emph{Comptes Rendus Mathematique}, 350\penalty0 (5-6):\penalty0 313--318, 2012.

\bibitem[Evtushenko and Posypkin(2014)]{evtushenko2014deterministic}
Yu~G Evtushenko and MA~Posypkin.
\newblock A deterministic algorithm for global multi-objective optimization.
\newblock \emph{Optimization Methods and Software}, 29\penalty0 (5):\penalty0 1005--1019, 2014.

\bibitem[Fliege and Svaiter(2000)]{fliege2000steepest}
J{\"o}rg Fliege and Benar~Fux Svaiter.
\newblock Steepest descent methods for multicriteria optimization.
\newblock \emph{Mathematical methods of operations research}, 51:\penalty0 479--494, 2000.

\bibitem[Fliege et~al.(2019)Fliege, Vaz, and Vicente]{fliege2019complexity}
J{\"o}rg Fliege, A~Ismael~F Vaz, and Lu{\'\i}s~Nunes Vicente.
\newblock Complexity of gradient descent for multiobjective optimization.
\newblock \emph{Optimization Methods and Software}, 34\penalty0 (5):\penalty0 949--959, 2019.

\bibitem[Franzmeyer et~al.(2022)Franzmeyer, Malinowski, and Henriques]{franzmeyer2022learning}
Tim Franzmeyer, Mateusz Malinowski, and João~F. Henriques.
\newblock Learning altruistic behaviours in reinforcement learning without external rewards, 2022.

\bibitem[Hu et~al.(2024)Hu, Xian, Wu, Fan, Yin, and Zhao]{hu2024revisiting}
Yuzheng Hu, Ruicheng Xian, Qilong Wu, Qiuling Fan, Lang Yin, and Han Zhao.
\newblock Revisiting scalarization in multi-task learning: A theoretical perspective.
\newblock \emph{Advances in Neural Information Processing Systems}, 36, 2024.

\bibitem[Kingma and Ba(2014)]{kingma2014adam}
Diederik~P Kingma and Jimmy Ba.
\newblock Adam: A method for stochastic optimization.
\newblock \emph{arXiv preprint arXiv:1412.6980}, 2014.

\bibitem[Kuba et~al.(2021)Kuba, Chen, Wen, Wen, Sun, Wang, and Yang]{kuba2021trust}
Jakub~Grudzien Kuba, Ruiqing Chen, Muning Wen, Ying Wen, Fanglei Sun, Jun Wang, and Yaodong Yang.
\newblock Trust region policy optimisation in multi-agent reinforcement learning.
\newblock \emph{arXiv preprint arXiv:2109.11251}, 2021.

\bibitem[Leonardos et~al.(2021)Leonardos, Overman, Panageas, and Piliouras]{leonardos2021global}
Stefanos Leonardos, Will Overman, Ioannis Panageas, and Georgios Piliouras.
\newblock Global convergence of multi-agent policy gradient in markov potential games.
\newblock \emph{arXiv preprint arXiv:2106.01969}, 2021.

\bibitem[Lin et~al.(2019)Lin, Zhen, Li, Zhang, and Kwong]{lin2019pareto}
Xi~Lin, Hui-Ling Zhen, Zhenhua Li, Qing-Fu Zhang, and Sam Kwong.
\newblock Pareto multi-task learning.
\newblock \emph{Advances in neural information processing systems}, 32, 2019.


\bibitem[Lowe et~al.(2017)Lowe, Wu, Tamar, Harb, Pieter~Abbeel, and Mordatch]{lowe2017multi}
Ryan Lowe, Yi~I Wu, Aviv Tamar, Jean Harb, OpenAI Pieter~Abbeel, and Igor Mordatch.
\newblock Multi-agent actor-critic for mixed cooperative-competitive environments.
\newblock \emph{Advances in neural information processing systems}, 30, 2017.


\bibitem[Momma et~al.(2022)Momma, Dong, and Liu]{momma2022multi}
Michinari Momma, Chaosheng Dong, and Jia Liu.
\newblock A multi-objective/multi-task learning framework induced by pareto stationarity.
\newblock In \emph{International Conference on Machine Learning}, pages 15895--15907. PMLR, 2022.

\bibitem[Papoudakis et~al.(2020)Papoudakis, Christianos, Sch{\"a}fer, and Albrecht]{papoudakis2020benchmarking}
Georgios Papoudakis, Filippos Christianos, Lukas Sch{\"a}fer, and Stefano~V Albrecht.
\newblock Benchmarking multi-agent deep reinforcement learning algorithms in cooperative tasks.
\newblock \emph{arXiv preprint arXiv:2006.07869}, 2020.

\bibitem[Rashid et~al.(2020)Rashid, Samvelyan, De~Witt, Farquhar, Foerster, and Whiteson]{rashid2020monotonic}
Tabish Rashid, Mikayel Samvelyan, Christian~Schroeder De~Witt, Gregory Farquhar, Jakob Foerster, and Shimon Whiteson.
\newblock Monotonic value function factorisation for deep multi-agent reinforcement learning.
\newblock \emph{Journal of Machine Learning Research}, 21\penalty0 (178):\penalty0 1--51, 2020.

\bibitem[Roy et~al.(2023)Roy, So, and Ma]{roy2023optimization}
Abhishek Roy, Geelon So, and Yi-An Ma.
\newblock Optimization on pareto sets: On a theory of multi-objective optimization.
\newblock \emph{arXiv preprint arXiv:2308.02145}, 2023.

\bibitem[Sener and Koltun(2018)]{sener2018multi}
Ozan Sener and Vladlen Koltun.
\newblock Multi-task learning as multi-objective optimization.
\newblock \emph{Advances in neural information processing systems}, 31, 2018.

\bibitem[Silver et~al.(2017)Silver, Hubert, Schrittwieser, Antonoglou, Lai, Guez, Lanctot, Sifre, Kumaran, Graepel, et~al.]{silver2017mastering}
David Silver, Thomas Hubert, Julian Schrittwieser, Ioannis Antonoglou, Matthew Lai, Arthur Guez, Marc Lanctot, Laurent Sifre, Dharshan Kumaran, Thore Graepel, et~al.
\newblock Mastering chess and shogi by self-play with a general reinforcement learning algorithm.
\newblock \emph{arXiv preprint arXiv:1712.01815}, 2017.

\bibitem[Sun et~al.(2022)Sun, Devlin, Beck, Hofmann, and Whiteson]{sun2022trust}
Mingfei Sun, Sam Devlin, Jacob Beck, Katja Hofmann, and Shimon Whiteson.
\newblock Trust region bounds for decentralized ppo under non-stationarity.
\newblock \emph{arXiv preprint arXiv:2202.00082}, 2022.

\bibitem[Sutton and Barto(2018)]{sutton2018reinforcement}
Richard~S Sutton and Andrew~G Barto.
\newblock \emph{Reinforcement learning: An introduction}.
\newblock MIT press, 2018.

\bibitem[Tampuu et~al.(2017)Tampuu, Matiisen, Kodelja, Kuzovkin, Korjus, Aru, Aru, and Vicente]{tampuu2017multiagent}
Ardi Tampuu, Tambet Matiisen, Dorian Kodelja, Ilya Kuzovkin, Kristjan Korjus, Juhan Aru, Jaan Aru, and Raul Vicente.
\newblock Multiagent cooperation and competition with deep reinforcement learning.
\newblock \emph{PloS one}, 12\penalty0 (4):\penalty0 e0172395, 2017.

\bibitem[Tan(1993)]{tan1993multi}
Ming Tan.
\newblock Multi-agent reinforcement learning: Independent vs. cooperative agents.
\newblock In \emph{Proceedings of the tenth international conference on machine learning}, pages 330--337, 1993.

\bibitem[Xiao et~al.(2024)Xiao, Ban, and Ji]{xiao2024direction}
Peiyao Xiao, Hao Ban, and Kaiyi Ji.
\newblock Direction-oriented multi-objective learning: Simple and provable stochastic algorithms.
\newblock \emph{Advances in Neural Information Processing Systems}, 36, 2024.

\bibitem[Yu et~al.(2022)Yu, Velu, Vinitsky, Gao, Wang, Bayen, and Wu]{yu2022surprising}
Chao Yu, Akash Velu, Eugene Vinitsky, Jiaxuan Gao, Yu~Wang, Alexandre Bayen, and Yi~Wu.
\newblock The surprising effectiveness of ppo in cooperative multi-agent games.
\newblock \emph{Advances in Neural Information Processing Systems}, 35:\penalty0 24611--24624, 2022.

\bibitem[Yu et~al.(2020)Yu, Kumar, Gupta, Levine, Hausman, and Finn]{yu2020gradient}
Tianhe Yu, Saurabh Kumar, Abhishek Gupta, Sergey Levine, Karol Hausman, and Chelsea Finn.
\newblock Gradient surgery for multi-task learning.
\newblock \emph{Advances in Neural Information Processing Systems}, 33:\penalty0 5824--5836, 2020.

\bibitem[Zeng et~al.(2019)Zeng, Dai, and Huang]{zeng2019convergence}
Liaoyuan Zeng, Yuhong Dai, and Yakui Huang.
\newblock Convergence rate of gradient descent method for multi-objective optimization.
\newblock \emph{Journal of Computational Mathematics}, 37\penalty0 (5):\penalty0 689, 2019.

\bibitem[Zhao et~al.(2023)Zhao, Yang, Wang, and Lee]{zhao2023local}
Yulai Zhao, Zhuoran Yang, Zhaoran Wang, and Jason~D Lee.
\newblock Local optimization achieves global optimality in multi-agent reinforcement learning.
\newblock \emph{arXiv preprint arXiv:2305.04819}, 2023.

\bibitem[Zhou et~al.(2022)Zhou, Zhang, Jiang, Zhong, Gu, and Zhu]{zhou2022convergence}
Shiji Zhou, Wenpeng Zhang, Jiyan Jiang, Wenliang Zhong, Jinjie Gu, and Wenwu Zhu.
\newblock On the convergence of stochastic multi-objective gradient manipulation and beyond.
\newblock \emph{Advances in Neural Information Processing Systems}, 35:\penalty0 38103--38115, 2022.

\end{thebibliography}

\end{document}